\documentclass[twocolumn]{article}

\usepackage{hyperref}
\usepackage{lipsum,graphicx,multicol}

\usepackage[utf8]{inputenc}
\usepackage{amsmath,amsthm,amssymb,amsfonts}
\usepackage[round]{natbib}
\usepackage{microtype}
\usepackage{graphicx,subfig}
\usepackage{algorithm,algorithmic}
\usepackage{booktabs,multirow}
\usepackage{fullpage}

\newcommand{\x}{\mathbf{x}}
\newcommand{\z}{\mathbf{z}}

\renewcommand{\u}{\mathbf{u}}
\newcommand{\y}{\mathbf{y}}

\newcommand{\f}{\mathbf{f}}
\newcommand{\g}{ {\mathbf{g} }}

\newcommand{\K}{\mathbf{K}}
\newcommand{\E}{\mathbb{E}}

\newcommand{\bR}{\mathbf{R}}
\newcommand{\Q}{\mathbf{Q}}
\newcommand{\Qf}{\mathbf{Q}_{\mathbf{f}}}
\newcommand{\C}{\mathbf{C}}
\newcommand{\U}{\mathbf{U}}
\newcommand{\Mf}{\mathbf{M}_{\mathbf{f}}}
\newcommand{\R}{\mathbb{R}}
\newcommand{\GP}{\mathcal{GP}}

\newcommand{\0}{\mathbf{0}}
\newcommand{\N}{\mathcal{N}}
\newcommand{\X}{\mathbf{X}}

\newcommand{\bS}{{\boldsymbol{\Sigma}}}

\newcommand{\bmu}{{\boldsymbol{\mu}}}

\newcommand{\Z}{\mathbf{Z}}

\newcommand{\XT}{{\mathbf{X}_T}}

\newcommand{\Zg}{{\mathbf{Z}_g}}
\newcommand{\Zf}{{\mathbf{Z}_\mathbf{f}}}
\newcommand{\Zft}{{\mathbf{Z}_\mathbf{f}^t}}
\newcommand{\Zfs}{{\mathbf{Z}_\mathbf{f}^s}}

\newcommand{\Uf}{{\mathbf{U}_\mathbf{f}}}

\newcommand{\ug}{{\mathbf{u}_g}}

\newcommand{\uf}{{\mathbf{u}_\mathbf{f}}}

\newcommand{\Sf}{{\mathbf{S}_\mathbf{f}}}

\newcommand{\mg}{{\mathbf{m}_g}}
\newcommand{\Sg}{{\mathbf{S}_g}}

\DeclareMathOperator{\KL}{\textsc{kl}}

\DeclareMathOperator{\cov}{\textbf{cov}}

\title{Deep learning with differential Gaussian process flows}

\author{
{\bf Pashupati Hegde, Markus Heinonen, Harri Lähdesmäki, Samuel Kaski} 
 \\ Helsinki Institute for Information Technology HIIT \\ Department of Computer Science, Aalto University }

\begin{document}

\maketitle
\begin{abstract}
We propose a novel deep learning paradigm of differential flows that learn a stochastic differential equation transformations of inputs prior to a standard classification or regression function. The key property of differential Gaussian processes is the warping of inputs through infinitely deep, but infinitesimal, differential fields, that generalise discrete layers into a dynamical system. We demonstrate state-of-the-art results that exceed the performance of deep Gaussian processes and neural networks.
\end{abstract}

\section{INTRODUCTION}

Gaussian processes are a family of flexible kernel function distributions \citep{rasmussen2006}. The capacity of kernel models is inherently determined by the function space induced by the choice of the kernel, where standard stationary kernels lead to models that underperform in practice. Shallow -- or single -- Gaussian processes are often suboptimal since flexible kernels that would account for the non-stationary and long-range connections of the data are difficult to design and infer. Such models have been proposed by introducing non-stationary kernels \citep{tolvanen2014expectation,heinonen2016}, kernel compositions \citep{duvenaud2011,sun2018}, spectral kernels \citep{wilson2013,remes2017}, or by applying input-warpings \citep{snoek2014} or output-warpings \citep{snelson2004,lazaro2012}. Recently, \citet{wilson2016} proposed to transform the inputs with a neural network prior to a Gaussian process model. The new neural input representation can extract high-level patterns and features, however, it employs rich neural networks that require careful design and optimization.

Deep Gaussian processes elevate the performance of Gaussian processes by mapping the inputs through multiple Gaussian process 'layers' \citep{damianou2013,salimbeni2017}, or as a network of GP nodes \citep{duvenaud2011,wilson2012,sun2018}. However, deep GP's result in degenerate models if the individual GP's are not invertible, which  limits their capacity \citep{duvenaud2014}.

In this paper we propose a novel paradigm of learning continuous-time transformations or \emph{flows} of the data instead of learning a discrete sequence of layers. We apply stochastic differential equation systems in the original data space to transform the inputs before a classification or regression layer. The transformation flow consists of an infinite path of infinitesimal steps.
This approach turns the focus from learning iterative function mappings to learning input representations in the original feature space, avoiding learning new feature spaces.

Our experiments show state-of-the-art prediction performance on a number of benchmark datasets on classification and regression. The performance of the proposed model exceeds that of competing Bayesian approaches, including deep Gaussian processes. 

\section{BACKGROUND}

We begin by summarising useful background of Gaussian processes and continuous-time dynamicals models. 

\subsection{Gaussian processes}

Gaussian processes (GP) are a family of Bayesian models that characterise distributions of functions \citep{rasmussen2006}. A Gaussian process prior on  a function $f(\x)$ over vector inputs $\x \in \R^D$,
\begin{align}
f(\x) &\sim \GP( 0, K(\x,\x')),
\end{align}
defines a prior distribution over function values $f(\x)$ whose mean and covariances are
\begin{align}
\E[ f(\x)] &= 0 \\
\cov[ f(\x), f(\x')] &= K(\x,\x').
\end{align}
A GP prior defines that for any collection of $N$ inputs, $\X = (\x_1, \ldots, \x_N)^T$, the corresponding function values $\f = ( f(\x_1), \ldots, f(\x_N))^T \in \R^N$ are coupled to following the multivariate normal distribution   
\begin{align}
\f \sim \N(\0, \K),
\end{align}
where $\K = (K(\x_i, \x_j))_{i,j=1}^N \in \R^{N \times N}$ is the kernel matrix. The key property of GP's is that output predictions $f(\x)$ and $f(\x')$ correlate depending on how similar are their inputs $\x$ and $\x'$, as measured by the kernel $K(\x,\x') \in \R$. 

We consider sparse Gaussian process functions by \emph{augmenting} the Gaussian process with a small number $M$ of \emph{inducing} `landmark' variables $u = f(\z)$ \citep{snelson2006sparse}. We condition the GP prior with the inducing variables $\u = (u_1, \ldots, u_M)^T \in \R^M$ and $\Z = (\z_1, \ldots, \z_M)^T$ to obtain the GP posterior predictions at data points
\begin{align}
    \f | \u ; \Z &\sim \N( \Q \u, \K_{\X \X} - \Q \K_{\Z \Z} \Q^T) \label{eq:interp} \\
    \u &\sim \N( \0, \K_{\Z \Z}), \label{eq:pu}
\end{align}
where $\Q = \K_{\X \Z} \K_{\Z \Z}^{-1}$, and where $\K_{\X \X} \in \R^{N \times N}$ is the kernel between observed image pairs $\X \times \X$, the kernel $\K_{\X \Z} \in \R^{N \times M}$ is between observed images $\X$ and inducing images $\Z$, and kernel $\K_{\Z \Z} \in \R^{M \times M}$ is between inducing images $\Z \times \Z$. The inference problem of sparse Gaussian processes is to learn the parameters $\theta$ of the kernel (such as the lengthscale), and the conditioning inducing variables $\u,\Z$.

\subsection{Stochastic differential equations}

\begin{figure}[t]
    \includegraphics[width=\columnwidth]{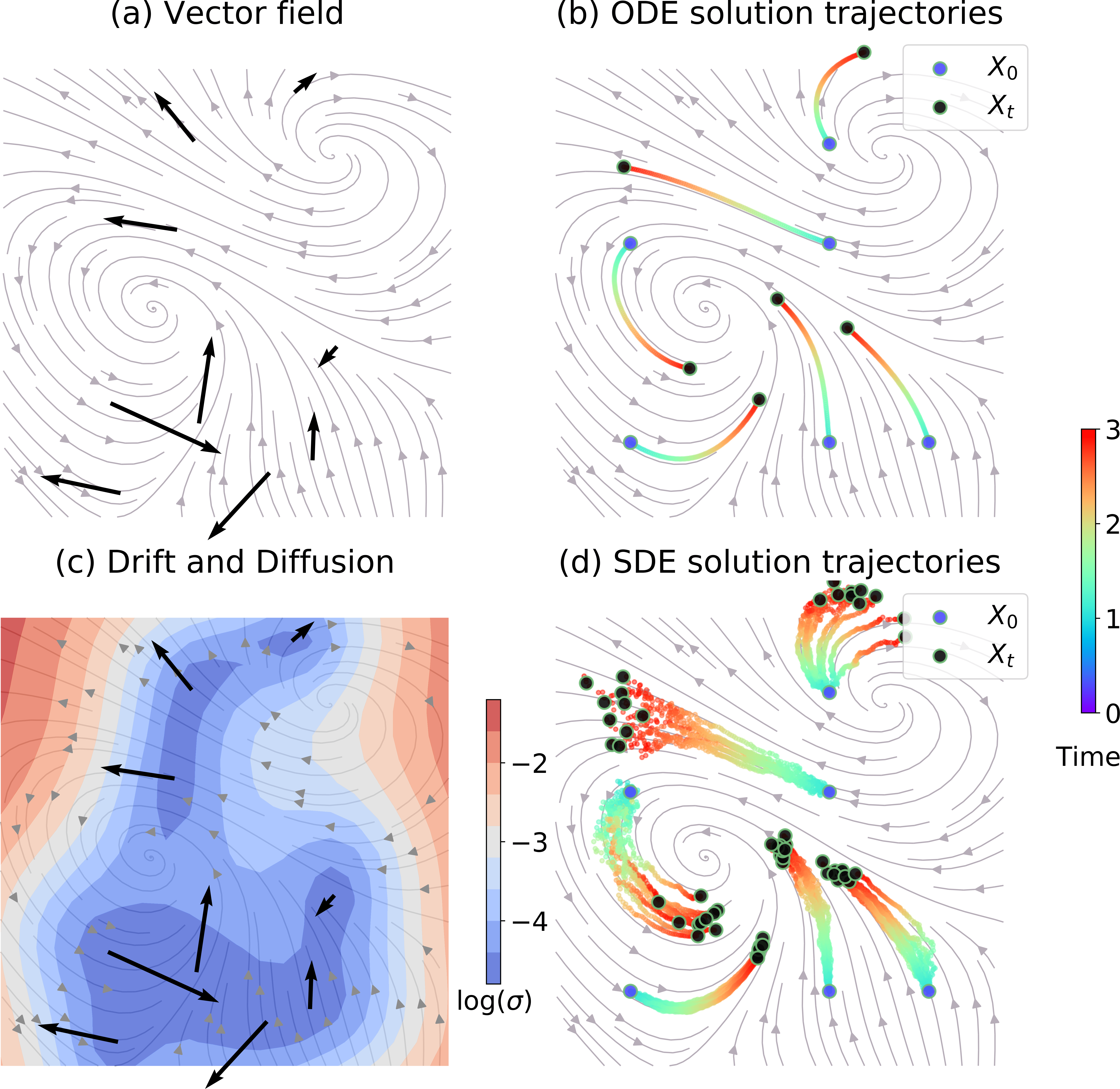}
    \caption{An example vector field defined by the inducing vectors \textbf{(a)} results in the ODE flow solutions \textbf{(b)} of a 2D system. Including the colored Wiener diffusion \textbf{(c)} leads to SDE trajectory distributions \textbf{(d)}.}
    \label{fig:odesde}
\end{figure}

Stochastic differential equations (SDEs) are an effective formalism for modelling continuous-time systems with underlying stochastic dynamics, with wide range of applications \citep{friedrich2011approaching}. We consider multivariate continuous-time systems governed by a Markov process $\x_t$ described by SDE dynamics
\begin{align} 
d\x_t = \bmu(\x_t)dt + \sqrt{\bS(\x_t)} dW_t, \label{eq:dx}
\end{align}
where $\x_t \in \R^D$ is the state vector of a $D$-dimensional dynamical system at continuous time $t \in \R$, $\bmu( \x_t ) \in \R^D$ is a deterministic state evolution vector field, $\sqrt{\Sigma(\x_t)} \in \R^{D \times D}$ is the diffusion matrix field of the stochastic multivariate Wiener process $W_t \in \R^D$. The $\sqrt{\Sigma(\x_t)}$ is the square root matrix of a covariance matrix $\Sigma(\x_t)$, where we assume $\Sigma(\x_t) = \sqrt{\Sigma(\x_t)} \sqrt{\Sigma(\x_t)}$ holds. A Wiener process has zero initial state $W_0 = \0$, and independent, Gaussian increments $W_{t+s} - W_{t} \sim \N(\0, s I_D)$ over time with standard deviation $\sqrt{s} I_D$ (See Figure \ref{fig:odesde}).

\begin{figure*}[t]
\centering
    \subfloat[Sparse GP]{\includegraphics[width=1.0in]{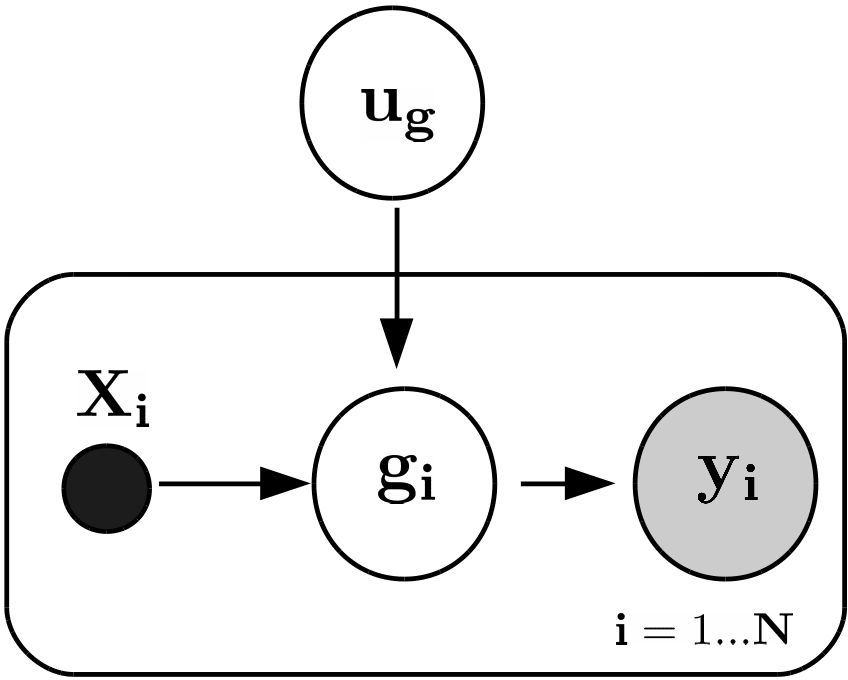}} \hspace{3mm}
    \subfloat[Deep GP]{\includegraphics[width=2.8in]{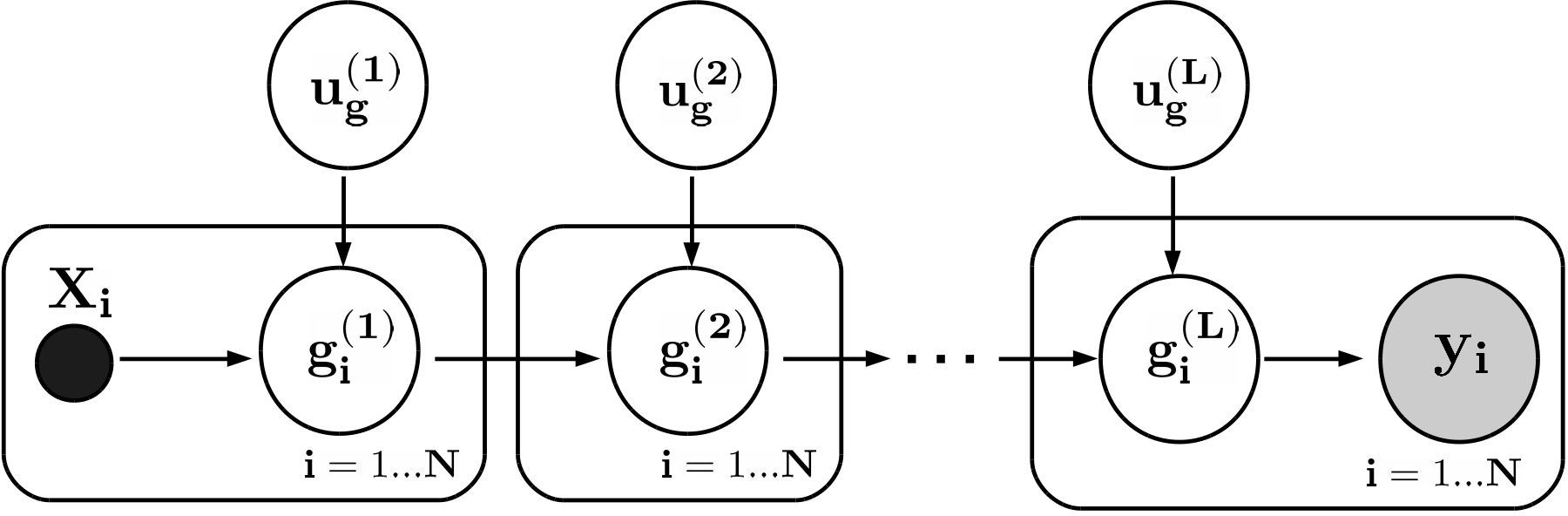}} \hspace{3mm}
    \subfloat[Differentially deep GP]{\includegraphics[width=2.2in]{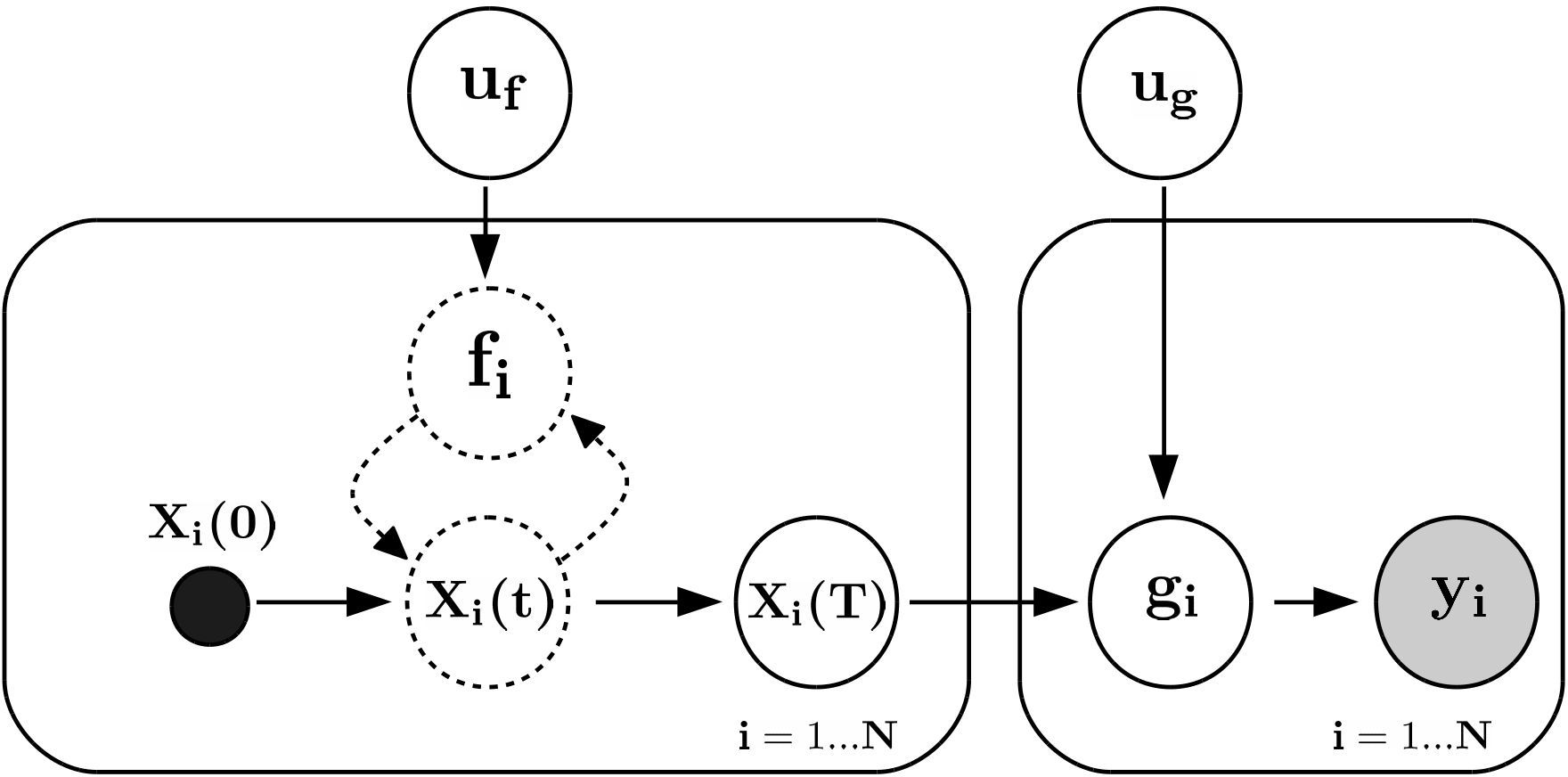}}
    \caption{The sparse Gaussian processes uncouples the observations through global inducing variables $\mathbf{u}_g$ \textbf{(a)}. Deep Gaussian process  is a hierarchical model with a nested composition of Gaussian processes introducing \textit{layer dependency}. Layer-specific inducing variables $\mathbf{u}_g^{(\ell)}$ introduce \textit{conditional independence} between function values $\mathbf{g}_i^{(\ell)}$ within each layer \textbf{(b)}. In our formulation deepness is introduced as a \textit{temporal dependency across} states $\x_i(t)$ (indicated by dashed line) with a GP prior over their differential function value $\f_i$  \textbf{(c)}. Global inducing variables $\uf$ can be used to introduce \textit{conditional independence between} differential function values at a particular time point.}
    \label{fig:plate}
\end{figure*}

The SDE system \eqref{eq:dx} transforms states $\x_t$ forward in continuous time by the deterministic \emph{drift} function $\bmu : \R^D \rightarrow \R^D$, while the \emph{diffusion} $\bS : \R^D \rightarrow \R^{D \times D}$ is the scale of the random Brownian motion $W_t$ that scatter the state $\x_t$ with random fluctuations. The state solutions of an SDE are given by the stochastic It\^o integral \citep{oksendal2014}
\begin{align}
\x_t &= \x_0 + \int_0^t \bmu(\x_\tau) d\tau + \int_0^t  \sqrt{\bS(\x_\tau)} d W_\tau, \label{eq:ito}
\end{align}
where we integrate the system state from an initial state $\x_0$ for time $t$ forward, and where $\tau$ is an auxiliary time variable. SDEs produce continuous, but non-smooth trajectories $\x_{0 : t}$ over time due to the non-differentiable Brownian motion. This causes the SDE system to not have a time derivative $\frac{d\x_t}{dt}$, but the  stochastic It\^o integral \eqref{eq:ito} can still be defined.

The only non-deterministic part of the solution \eqref{eq:ito} is the Brownian motion $W_\tau$, whose random realisations generate path realisations $\x_{0 : t}$ that induce state distributions 
\begin{align}
    \x_t \sim p_t(\x ; \bmu, \bS, \x_0) \label{eq:sd}
\end{align}
at any instant $t$, given the drift $\bmu$ and diffusion $\bS$ from initial state $\x_0$. The state distribution is the solution to the Fokker-Planck-Kolmogorov (FPK) partial differential equation, which is intractable for general non-linear drift and diffusion.

In practise the Euler-Maruyama (EM) numerical solver can be used to simulate trajectory samples from the state distribution \citep{yildiz2018} (See Figure \ref{fig:odesde}d). We assume a fixed time discretisation $t_1, \ldots, t_N$ with $\Delta t = t_N/N$ being the time window \citep{higham2001}. The EM method at $t_k$ is
\begin{align}
\x_{k+1} &= \x_k + \bmu(\x_k) \Delta t + \sqrt{\Sigma(\x_k)} \Delta W_k, \label{eq:em}
\end{align}
where $\Delta W_k = W_{k+1} - W_k \sim \N(\0, \Delta t I_D)$ with standard deviation $\sqrt{\Delta t}$. The EM increments $\Delta \x_k = \x_{k+1} - \x_k$ correspond to samples from a Gaussian
\begin{align}
    \Delta \x_k \sim \N(\bmu(\x_k) \Delta t, \Sigma(\x_k) \Delta t ). \label{eq:emdist}
\end{align}
Then, the full $N$ length path is determined from the $N$ realisations of the Wiener process, each of which is a $D$-dimensional. More efficient high-order approximations have also been developed \citep{kloeden1992numerical,lamba2006}.

SDE systems are often constructed by manually defining drift and diffusion functions to model specific systems in finance, biology, physics or in other domains \citep{friedrich2011approaching}. Recently, several works have proposed learning arbitrary drift and diffusion functions from data \citep{papaspiliopoulos2012,garcia2017,yildiz2018}.

\section{DEEP DIFFERENTIAL GAUSSIAN PROCESS}
\label{sec:flowgp}

In this paper we propose a paradigm of \emph{continuous-time deep learning}, where inputs $\x_i$ are not treated as constant, but are instead driven by an SDE system. We propose a continuous-time deep Gaussian process model through \emph{infinite}, infinitesimal differential compositions, denoted as DiffGP. In DiffGP, a Gaussian process warps or \emph{flows} an input $\x$ through an SDE system until a predefined time $T$, resulting in $\x(T)$, which is subsequently classified or regressed with a separate function. We apply the process to both train and test inputs. We impose GP priors on both the stochastic differential fields and the predictor function (See Figure \ref{fig:plate}). 
A key parameter of the differential GP model is the amount of simulation time $T$, which defines the length of flow and the capacity of the system, analogously to the number of layers in standard deep GPs or deep neural networks.

We assume a dataset of $N$ inputs $\X = (\x_1, \ldots, \x_N)^T \in \R^{N \times D}$ of $D$-dimensional vectors $\x_i \in \R^D$, and associated scalar outputs $\y = (y_1, \ldots, y_N)^T \in \R^N$ that can be continuous for a regression problem or categorical for classification, respectively. We redefine the inputs as temporal functions $\x : \mathcal{T} \rightarrow \R^D$ over time such that state paths $\x_t$ over time $t \in \mathcal{T} = \R_+$ emerge, where the observed inputs $\x_{i,t} \triangleq \x_{i,0}$ correspond to initial states $\x_{i,0}$ at time $0$. We classify or regress the final data points $\X_T = (\x_{1,T}, \ldots, \x_{N,T} )^T$ after $T$ time of an SDE flow with a predictor Gaussian process
\begin{align}
g( \x_T ) & \sim \GP( 0, K(\x_T, \x'_T )) \label{eq:diffgp}
\end{align}
to classify or regress the outputs $\y$. The framework reduces to a conventional Gaussian process with zero flow time $T=0$ (See Figure \ref{fig:plate}). 

The prediction depends on the final dataset $\X_T$ structure, determined by the SDE flow $d\x_t$ from the original data $\X$. We consider SDE flows of type
\begin{align}
    d\x_t = \bmu(\x_t) dt + \sqrt{\Sigma(\x_t)} dW_t \label{eq:sdeflow}
\end{align}
where
\begin{align}
\bmu(\x) &= \K_{\x \Zf} \K_{\Zf \Zf}^{-1} \mathrm{vec}(\Uf) \\
\bS(\x) &= \K_{\x \x} - \K_{\x \Zf} \K_{\Zf \Zf}^{-1} \K_{\Zf \x}
\end{align}
are the vector-valued Gaussian process posteriors conditioned on equalities between $\f(\z)$, inducing vectors $\Uf = (\u_1^f, \ldots, \u_M^f)^T$ and inducing states $\Zf = (\z_1^f, \ldots, \z_M^f)$. These choices of drift and diffusion correspond to an underlying GP
\begin{align}
    \f(\x) &\sim \GP( \0, K(\x,\x') ) \\
    \f(\x) | \Uf,\Zf  & \sim \N( \bmu(\x), \bS(\x) ) \label{eq:fuz}
\end{align}
where $K(\x,\x') \in \R^{D \times D}$ is a matrix-valued kernel of the vector field $\f(\x) \in \R^D$, and $\K_{\Zf \Zf} = (K(\z_i^f,\z_j^f))_{i,j=1}^M \in \R^{MD \times MD}$ block matrix of matrix-valued kernels (similarly for $\K_{\x \Zf}$). 

The \emph{vector field} $\f(\x)$ is now a GP with deterministic conditional mean $\bmu$ and covariance $\bS$ at every location $\x$ given the inducing variables. We encode the underlying GP field mean and covariance uncertainty into the drift and diffusion of the SDE flow \eqref{eq:sdeflow}. The Wiener process  $W_t$ of an SDE samples a new fluctuation from the covariance $\bS$ around the mean $\bmu$ at every instant $t$. This corresponds to an affine transformation
\begin{align}
    (\f(\x) - \bmu(\x)) \sqrt{\Delta t} + \bmu(\x) \Delta t &\sim \N(\bmu(\x) \Delta t, \Sigma(\x) \Delta t ),
\end{align}
which shows that samples from the vector field GP match the SDE Euler-Maruyama increment $\Delta \x_k$ distribution \eqref{eq:emdist}. The state distribution $p_T(\x ; \bmu, \bS, \x_0)$ can then be represented as $p(\x_T | \Uf) = \int p(\x_T | \f) p(\f | \Uf) d\f$ 
where $p(\x_T | \f)$ is an Dirac distribution of the end point of a single Euler-Maruyama simulated path, and where the vector field $p(\f)$ is marginalized along the Euler-Maruyama path.

Our model corresponds closely to the doubly-stochastic deep GP, where the Wiener process was replaced by random draws from the GP posterior $\varepsilon^l \cdot  \Sigma^{l}( \f^{l-1} )$ per layer $l$ \citep{salimbeni2017}. In our approach the continuous time $t$ corresponds to continuously indexed states, effectively allowing infinite layers that are infinitesimal.

\subsection{Spatio-temporal fields}
\label{spatiotemporal}

Earlier we assumed a global, time-independent vector field $\f(\x_t)$, which in the standard models would correspond to a single `layer' applied recurrently over time $t$. To extend the model capacity, we consider spatio-temporal vector fields $\f_t(\x) := \f(\x, t)$ that themselves evolve as a function of time, effectively applying a smoothly changing vector field `layer' at every instant $t$. We select a separable spatio-temporal kernel $K( (\x,t), (\x',t')) = K(\x,\x') k(t,t')$ that leads to an efficient Kronecker-factorised  \citep{stegle2011} spatio-temporal SDE flow
\begin{align}    
\f_t(\x) | (\Zfs,\Zft,\Uf)  &\sim \N(\bmu_t(\x), \bS_t(\x)) \\
 \bmu_t(\x) &= \C_{\x \Zf} \C_{\Zf \Zf}^{-1} \mathrm{vec}(\Uf) \\
 \bS_t(\x) &= \C_{\x \x} - \C_{\x \Zf} \C_{\Zf \Zf}^{-1} \C_{\Zf \x},
\end{align}
where $\C_{\x \x} = K_{\x \x} k_{t t}$, $\C_{\x \Z} = \K_{\x \Zfs} \otimes \K_{t \Zft}$ and $\C_{\Zf \Zf} = \K_{\Zfs \Zfs} \otimes \K_{\Zft \Zft}$, and where the \emph{spatial} inducing states are denoted by $\Zfs$ and the \emph{temporal} inducing times by $\Zft$. In practice we place usually only a few (e.g. 3) temporal inducing times equidistantly on the range $[0,T]$. This allows the vector field itself to curve smoothly throughout the SDE. We only have a single inducing matrix $\Uf$ for both spatial and temporal dimensions.

\begin{figure*}[t]
    \centering
    \includegraphics[width=0.8\textwidth]{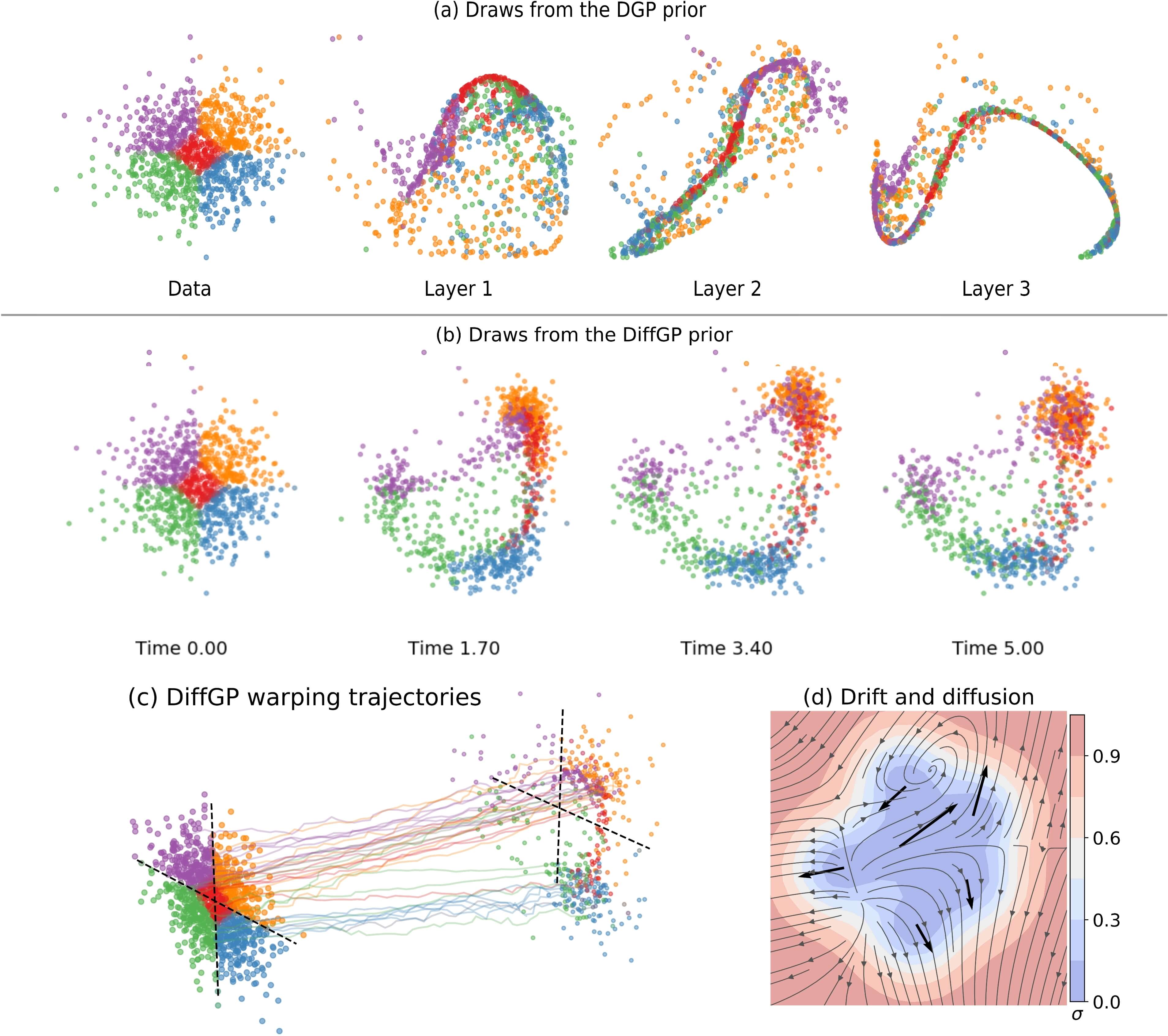}
    \caption{\textbf{(a)}Illustration of samples from a 2D deep Gaussian processes prior. DGP prior exhibits a pathology wherein representations in deeper layers concentrate on low-rank manifolds.\textbf{(b)} Samples from a differentially deep Gaussian processes prior result in rank-preserving representations.\textbf{(c)} The continuous-time nature of the warping trajectories results from smooth drift and structured diffusion \textbf{(d)}.}
    \label{fig:prior}
\end{figure*}

\subsection{Stochastic variational inference}

The differential Gaussian process is a combination of a conventional prediction GP $g(\cdot)$ with an SDE flow GP $\f(\cdot)$ fully parameterised by $\Z,\U$ as well as kernel parameters $\theta$. We turn to variational inference to estimate posterior approximations $q(\Uf)$ and $q(\u_g)$ for both models.

We start by augmenting the predictor function $g$ with $M$ inducing locations $\Zg = (\z_{g1}, \ldots, \z_{gM})$ with associated inducing function values $g(\z) = u$ in a vector $\ug = (u_{g1}, \ldots, u_{gM})^T \in \R^M$. We aim to learn the distribution of the inducing values $\u$, while learning point estimates of the inducing locations $\Z$, which we hence omit from the notation below. The prediction conditional distribution is \citep{titsias2009}
\begin{align}
p(\g | \ug, \XT) &= \N(\g | \Q_T \ug, \K_{\XT \XT} - \Q_T \K_{\Zg \Zg} \Q_T^T)  \label{eq:hu}  \\
    p(\ug ) &= \N(\ug | \0, \K_{\Zg \Zg}),  \label{eq:prioru}
\end{align}
where we denote $\Q_T = \K_{\XT \Zg} \K_{\Zg \Zg}^{-1}$.

The joint density of a single path and prediction of the augmented system is
\begin{align}
    &p(\y, \g, \ug, \XT, \f,\Uf ) \\
    &= \underbrace{p(\y | \g)}_{\text{likelihood}} \underbrace{p(\g | \ug, \XT)  p(\ug )}_{\text{GP prior of $g(\x)$}} \underbrace{p(\XT|\f )}_{\text{SDE}} \underbrace{p(\f | \Uf) p(\Uf)}_{\text{GP prior of $\f(\x)$}}. \notag
\end{align}
The joint distribution contains the likelihood term, the two GP priors, and the SDE term $p(\XT | \f)$ representing the Euler-Maruyama paths of the dataset. 
The inducing vector field prior follows

\begin{align}
    p(\Uf) &= \prod_{d=1}^D \N( \mathbf{u}_{\f d} | \0, \K_{\Zf_d \Zf_d}), \label{eq:priorf}
\end{align}
where $\mathbf{u}_{\f d} = (\u_1^f(d)^T, \ldots, \u_M^f(d))$ and $\Zf_d = (\z_1^f(d), \ldots, \z_M^f(d))^T$.

We consider optimizing the marginal likelihood
\begin{align}
\log p(\y) &= \log \E_{ p(\g|\XT)p(\XT) } p(\y | \g), \label{eq:mll}
\end{align}
where the $p(\g|\XT)$ is a Gaussian process predictive distribution, and the state distribution $p(\XT)$ marginalizes the trajectories,
\begin{align}
 p(\XT) &= \iint p(\XT | \f ) p(\f | \Uf) p(\Uf) d\f d\Uf,
\end{align}
with no tractable solution. 

We follow stochastic variational inference (SVI) by \citet{hensman2015}, where standard variational inference \citep{blei2016} is applied to find a lower bound of the marginal log likelihood, or in other words model evidence. In particular, a variational lower bound for the evidence \eqref{eq:mll} without the state distributions has already been considered by \citet{hensman2015}, which tackles both problems of cubic complexity $O(N^3)$ and marginalization of non-Gaussian likelihoods. We propose to include the state distributions by simulating Monte Carlo state trajectories.

We propose a complete variational posterior approximation over both $\f$ and $g$,
\begin{align}
q(\g, \ug, \XT, \f, \Uf) &= p(\g | \ug,\XT) q(\ug) \\ \nonumber
&\qquad \cdot \: p(\XT|\f) p(\f | \Uf) q(\Uf) \\
q(\ug) &= \N(\ug | \mg, \Sg) \\ 
q(\Uf) &= \prod_{d=1}^D \N( \mathbf{u}_{\f d} | \mathbf{m}_{\f d}, \mathbf{S}_{\f d}),
\end{align}
where $\Mf = ( \mathbf{m}_{\mathbf{f} 1}, \ldots, \mathbf{m}_{\mathbf{f} D})$ and $\Sf = (\mathbf{S}_{\f 1}, \ldots, \mathbf{S}_{\f D})$ collect the dimension-wise inducing parameters. We continue by marginalizing out inducing variables $\ug$ and $\Uf$ from the above joint distribution, arriving at the joint variational posterior
\begin{align}
q(\g, \XT, \f) &= q(\g | \XT) p(\XT | \f) q(\f), \label{eq:postrj}
\end{align}
where
\begin{align}
    q(\g | \XT) &= \int p( \g | \ug, \XT) q(\ug) d \ug \label{eq:postrg} \\
    & \hspace{-1cm}= \N( \g |\Q_T \mg, \K_{\XT \XT} + \Q_T(\Sg - \K_{\Zg \Zg}) \Q_T^T) \\
    q(\f) &= \int p( \f | \Uf) q(\Uf) d \Uf = \N( \f |\bmu_q,\bS_q) \label{eq:postrf} \\
    \bmu_q &= \Qf \mathrm{vec} (\Mf) \\
    \bS_q  &= \K_{\X \X} + \Qf (\Sf - \K_{\Zf \Zf}) \Qf^T,
\end{align}
where $\Qf = \K_{\X \Zf} \K_{\Zf \Zf}^{-1}$. We plug the derived variational posterior drift  $\bmu_q$ and diffusion $\bS_q$ estimates to the final  \emph{variational SDE flow}
\begin{align}
    d\x_t &= \bmu_q(\x_t) dt + \sqrt{\bS_q(\x_t)} dW_t, \label{eq:postr_sde}
\end{align}
which conveniently encodes the variational approximation of the vector field $\f$.

Now the lower bound for our differential deep GP model can be written as (detailed derivation is provided in the appendix)
\begin{align}
    \log p(\y) &\ge \sum_{i=1}^N \bigg\{ \frac{1}{S} \sum_{s=1}^S \underbrace{\E_{q( \g | \x_{i,T}^{(s)}) } \log p(y_i | g_i)}_{\text{variational expected likelihood}} \notag \\
    &\hspace{-10mm} - \underbrace{\KL[ q(\ug) || p(\ug)]}_{\text{prior divergence of $g(x)$}} - \underbrace{\KL[ q(\Uf) || p(\Uf)]}_{\text{prior divergence of $\f(x)$}} \bigg\}, \label{eq:elbo3}
\end{align}
which factorises over both data and SDE paths with unbiased samples $\x_{i,T}^{(s)} \sim p_T( \x ; \bmu_q, \bS_q, \x_i)$ by numerically solving the variational SDE \eqref{eq:postr_sde} using the Euler-Maruyama method.

For likelihoods such as Gaussian for regression problems, we can further marginalize $\g$ from the lowerbound as shown by \citet{hensman2013}. For other intractable likelihoods, numerical integration techniques such as Gauss-Hermite quadrature method can be used \citep{hensman2015}.

\subsection{Rank pathologies in deep models}
A deep Gaussian process $\f^L( \cdots \f^2(\f^1(\x)))$ is a composition of $L$ Gaussian process layers $\f^l(\x)$ \citep{damianou2013}. These models typically lead to degenerate covariances, where each layer in the composition reduces the rank or degrees of freedom of the system \citep{duvenaud2014}. In practice the rank reduces via successive layers mapping inputs to identical values (See Figure \ref{fig:prior}a), effectively merging inputs and resulting in a reduced-rank covariance matrix with repeated rows and columns. To counter this pathology \citet{salimbeni2017} proposed pseudo-monotonic deep GPs by using identity mean function in all intermediate GP layers.

Unlike the earlier approaches, our model does not seem to suffer from this degeneracy. The DiffGP model warps the input space without seeking low-volume representations. In particular the SDE diffusion scatters the trajectories preventing both narrow manifolds and input merging. In practice, this results in a rank-preserving model (See Figure \ref{fig:prior}b-d). Therefore, we can use zero mean function for the Gaussian processes responsible for differential warpings.

\section{EXPERIMENTS}
\begin{table*}[th]
\resizebox{1.00\textwidth}{!}{
\begin{tabular}{ l r  cc cc cc cc }
\toprule
& & boston & energy & concrete & wine\_red & kin8mn & power & naval & protein \\
\cmidrule(lr){3-10}
& $N$ & 506 & 768 & 1,030 & 1,599 & 8,192 & 9,568 & 11,934 & 45,730  \\
& $D$ & 13 & 8 & 8 & 22 & 8 & 4 & 26 & 9  \\
\midrule
Linear && 4.24(0.16) & 2.88(0.05) & 10.54(0.13) & 0.65(0.01) & 0.20(0.00) & 4.51(0.03) & 0.01(0.00) & 5.21(0.02)  \\
\cmidrule(lr){1-10}
BNN & $L=2$          & 3.01(0.18) & 1.80(0.05) & 5.67(0.09) & 0.64(0.01) & 0.10(0.00) & 4.12(0.03) & 0.01(0.00) & 4.73(0.01)  \\
\cmidrule(lr){1-10}
\multirow{2}{*}{Sparse GP}& $M=100$  & 2.87(0.15) & 0.78(0.02) & 5.97(0.11) & 0.63(0.01) & 0.09(0.00) & 3.91(0.03) & \textbf{0.00}(0.00) & 4.43(0.03)  \\
& $M=500$  & 2.73(0.12) & \textbf{0.47}(0.02) & 5.53(0.12) & \textbf{0.62}(0.01) & 0.08(0.00) & 3.79(0.03) &\textbf{0.00}(0.00) & 4.10(0.03)  \\
\cmidrule(lr){1-10}
\multirow{4}{*}{\shortstack{Deep GP \\ $M=100$}} & $L=2$      & 2.90(0.17) & \textbf{0.47}(0.01) & 5.61(0.10) & 0.63(0.01) & \textbf{0.06}(0.00) & 3.79(0.03) & \textbf{0.00}(0.00) & 4.00(0.03)  \\
& $L=3$      & 2.93(0.16) & 0.48(0.01) & 5.64(0.10) & 0.63(0.01) & \textbf{0.06}(0.00) & 3.73(0.04) & \textbf{0.00}(0.00) & \textbf{3.81}(0.04)  \\
& $L=4$      & 2.90(0.15) & 0.48(0.01) & 5.68(0.10) & 0.63(0.01) & \textbf{0.06}(0.00) & 3.71(0.04) & \textbf{0.00}(0.00) & \textbf{3.74}(0.04)  \\
& $L=5$      & 2.92(0.17) & \textbf{0.47}(0.01) & 5.65(0.10) & 0.63(0.01) & \textbf{0.06}(0.00) & 3.68(0.03) & \textbf{0.00}(0.00) & \textbf{3.72}(0.04)  \\
\cmidrule(lr){1-10}
\multirow{5}{*}{\shortstack{DiffGP \\ $M=100$}}
& $T= 1.0$ &2.80(0.13) & 0.49(0.02) & \textbf{5.32}(0.10) & 0.63(0.01) & \textbf{0.06}(0.00) & 3.76(0.03) & \textbf{0.00}(0.00) & 4.04(0.04) \\ 
& $T= 2.0$ & \textbf{2.68}(0.10) & 0.48(0.02) & \textbf{4.96}(0.09) & 0.63(0.01) & \textbf{0.06}(0.00) & 3.72(0.03) & \textbf{0.00}(0.00) & 4.00(0.04) \\
& $T= 3.0$ & \textbf{2.69}(0.14) & \textbf{0.47}(0.02)& \textbf{4.76}(0.12) & 0.63(0.01) & \textbf{0.06}(0.00) & 3.68(0.03) & \textbf{0.00}(0.00) & 3.92(0.04)  \\
& $T= 4.0$ & \textbf{2.67}(0.13) & 0.49(0.02) & \textbf{4.65}(0.12) & 0.63(0.01) & \textbf{0.06}(0.00) & \textbf{3.66}(0.03) & \textbf{0.00}(0.00) & 3.89(0.04) \\
& $T= 5.0$ & \textbf{2.58}(0.12) & 0.50(0.02) & \textbf{4.56}(0.12) & 0.63(0.01) & \textbf{0.06}(0.00) & \textbf{3.65}(0.03) & \textbf{0.00}(0.00) & 3.87(0.04)  \\
\bottomrule
\end{tabular}
}
    \caption{Test RMSE values of 8 benchmark datasets (reproduced from Salimbeni \& Deisenroth 2017). Uses random 90\% / 10\% training and test splits, repeated 20 times. }
    \label{tab:results}
\end{table*}
We optimize the inducing vectors, inducing locations, kernel lengthscales and signal variance of both the SDE function $\f$ equation~\eqref{eq:sdeflow} and the predictor function $g(\x_T)$. We also optimize noise variance in problems with Gaussian likelihoods. The number of inducing points $M$ is manually chosen, where more inducing points tightens the variational approximation at the cost of additional computation. All parameters are jointly optimised against the evidence lower bound \eqref{eq:elbo3}. The gradients of the lower bound back-propagate through the prediction function $g(\x_T)$ and through the SDE system from $\x(T)$ back to initial values $\x(0)$. Gradients of an SDE system approximated by an EM method can be obtained with the autodiff differentiation of TensorFlow \citep{abadi2016}. The gradients of continuous-time systems follow from forward or reverse mode sensitivity equations \citep{kokotovic1967,raue2013,froechlich2017, yildiz2018}. We perform stochastic optimization with mini-batches and the Adam optimizer \citep{kingma2014adam} with a step size of 0.01. For numerical solutions of SDE, we use Euler-Maruyama solver with 20 time steps. Also, initializing parameters of  $g(\cdot)$ with values learned through SGP results in early convergence; we initialize DiffGP training with SGP results and a very weak warping field $\Uf \approx 0$ and kernel variance $\sigma_f^2 \approx 0.01$. We use diagonal approximation of the $\bS_q$. We also use GPflow \citep{GPflow2017}, a Gaussian processes framework built on TensorFlow in our implementation.

\subsection{Step function estimation}
\begin{figure*}[th]
    \centering
    \includegraphics[width=1.0\textwidth]{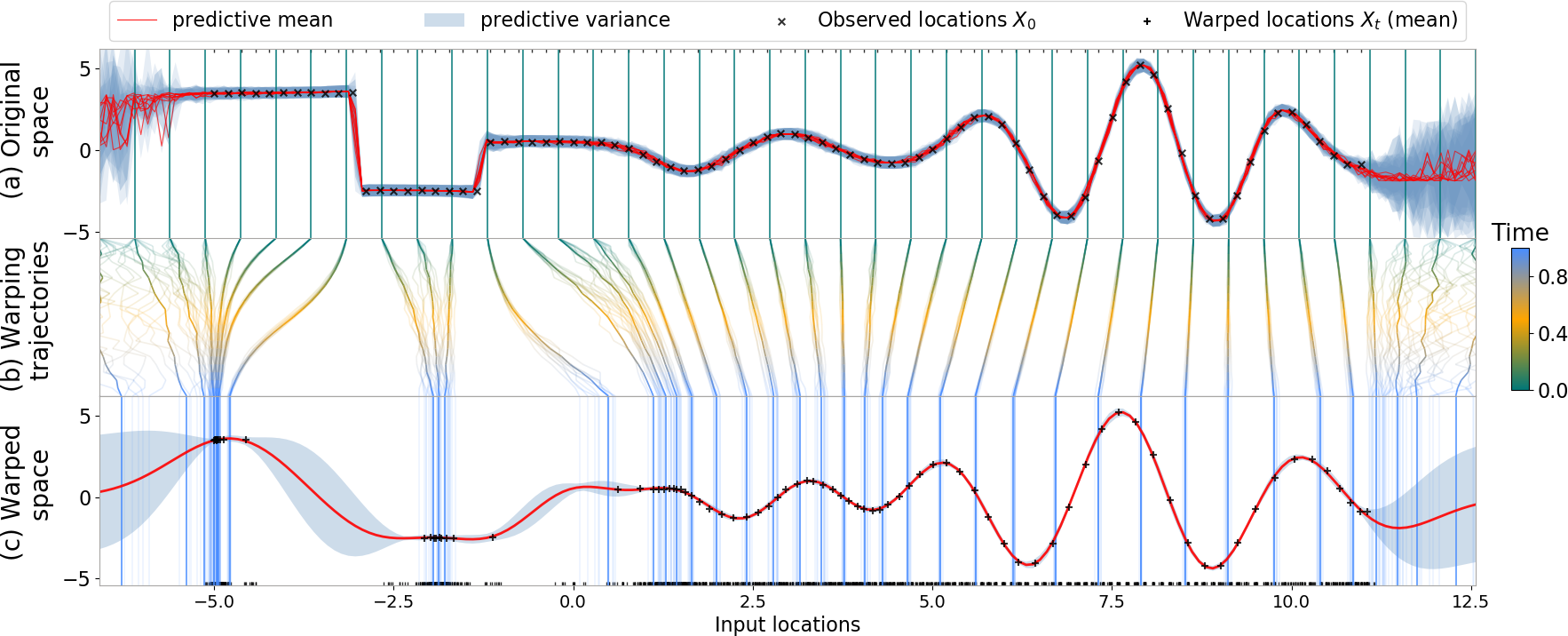}
    \caption{Step function estimation: Observed input space \textbf{(a)} 
    is transformed through stochastic continuous-time mappings \textbf{(b)} into a warped space \textbf{(c)}. The stationary Gaussian process in the warped space gives a smooth predictive distribution corresponding to a highly non-stationary predictions in the original observed space.} 
    \label{fig:step}
\end{figure*}

We begin by highlighting how the DiffGP estimates a signal with multiple highly non-stationary step functions. Figure \ref{fig:step} shows the univariate signal observations (top), the learned SDE flow (middle), and the resulting regression function on the end points $\X(t)$ (bottom). The DiffGP separates the regions around the step function such that the final regression function $g$ with a standard stationary Gaussian kernel can fit the transformed data $\X(t)$. The model then has learned the non-stationarities of the system with uncertainty in the signals being modelled by the inherent uncertainties arising from the diffusion.

\subsection{UCI regression benchmarks}
We compare our model on 8 regression benchmarks with the previously reported state-of-the-art results in \citep{salimbeni2017}. We test all the datasets on different flow time values from 1 to 5. We use the RBF kernel with ARD and 100 inducing points for both the differential Gaussian Process and the regression Gaussian Process. Each experiment is repeated 20 times with random 90\% / 10\% training and test splits. While testing, we compute predictive mean and predictive variance for each of the sample generated from \eqref{eq:postr_sde}, and compute the average of summary statistics (RMSE and log likelihood) over these samples. The mean and standard error of RMSE values are reported in Table 1.

On Boston, Concrete and Power datasets, where deep models show improvement over shallow models, our model outperforms previous best results of DGP. There is a small improvement by having a non-linear model on the Kin8mn dataset and our results match that of DGP. Energy and Wine are small datasets where single Gaussian Processes perform the best. As expected, both DiffGP and DGP recover the shallow model indicating no over-fitting. Regression task on the Protein dataset is aimed at predicting RMSD (Root Mean Squared Deviation) between modeled and native protein structures using 9 different properties of the modeled structures \citep{rana2015quality}. We suspect DGP particularly performs better than DiffGP in the task because of its capability to model long-range correlations.

\subsection{UCI classification benchmarks}

We perform binary classification experiments on large-scale HIGGS and SUSY datasets with a data size in the order of millions. We use the AUC as the performance measure and compare the results with the previously reported results using DGP \citep{salimbeni2017} and DNN \citep{baldi2014searching}. The classification task involves identifying processes that produce Higgs boson and super-symmetric particles using data from Monte Carlo simulations. Previously, deep learning methods based on neural networks have shown promising results on these tasks \citep{baldi2014searching}. On the HIGGS dataset, the proposed DiffGP model shows state-of-the-art (0.878) results, equal or even better than the earlier reported results using DGPs (0.877) and DNNs (0.876). On the SUSY dataset, we reach the performance of 4-hidden layer DGP (0.841) with non-temporal DiffGP (0.842). Considering the consistent improvement in the performance of DGP models with additional layers, we tried increasing the capacity of DiffGP model using the temporal extension proposed in Section \ref{spatiotemporal}. In particular, we used 100 spatial inducing vectors along with 3 temporal inducing vectors. The temporal DiffGP model gives an AUC of 0.878 on HIGGS and 0.846 on SUSY datasets matching the best reported results of DGP (see appendix for detailed comparison).

\subsection{Importance of flow time}
\begin{figure}[th]
    \centering
    \includegraphics[width=1.0\columnwidth]{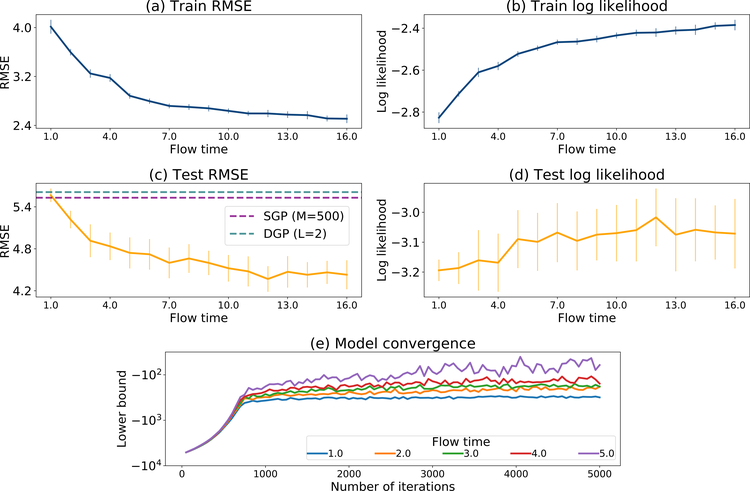}
    \caption{Concrete dataset: increasing the flow time variable $T$ improves the train and test errors \textbf{(a,c)} and likelihoods \textbf{(b,d)}. The horizontal line indicates GP and DGP2 performance. The model convergence indicates the improved capacity upon increased flow time \textbf{(e)}.}
    \label{fig:time}
\end{figure}

In this we experiment we study the SDE flow time parameter on Concrete dataset. Increasing integration time provides more warping flexibility to the SDE warping component. That is, with increase in the flow time, the SDE system can move observations further away from the initial state, however at the cost of exposing the state to more diffusion which acts as a principled regularization. Thus increasing time can lead to an increase in the model capacity without over-fitting. We empirically support this claim in the current experiment by fitting a regression model multiple times and maintaining same experimental setup, expect for the flow time. Figure \ref{fig:time} shows the variation in RMSE, log likelihood and the lower bound on marginal likelihood across different flow times. It can be seen that the improvement in the performance almost saturates near time = $10$.

\section{DISCUSSION}
We have proposed a novel paradigm, continuous-time Gaussian process deep learning. The proposed deferentially deep composition is a continuous-time approach wherein a Gaussian processes input locations are warped through stochastic and smooth differential equations. This results in a principled Bayesian approach with a smooth non-linear warping; the uncertainty through diffusion acts as a key regularizer. 

We empirically show excellent results in various regression and classification tasks. Also, DGP with the model specification as proposed by \citet{salimbeni2017}, uses a total of $\mathcal{O}(LDM)$ number of inducing parameters for the regression results, where $L$ is the number of layers, $D$ is the input dimension, $M$ is the number of inducing points for each latent GP. In contrast, with a smaller number of inducing parameters $\mathcal{O}(DM)$, we arrive at similar or even better results. 

The continuous-time deep model admits `decision-making paths', where we can explicitly follow the transformation applied to a data point $\x_i$. Analyzing these paths could lead to a better interpretable model. However, modeling in the input space without intermediate low-dimensional latent representations presents scalability issues. We leave scaling the approach to high dimensions as future work, while we also intend to explore new optimisation modes, such as SG-MCMC \citep{ma2015complete} or Stein inference \citep{liu2016stein} in the future.

\bibliographystyle{plainnat}

\clearpage
\onecolumn
\section*{APPENDIX}
\subsection*{A. Derivation of the stochastic variational inference}

The differential Gaussian process is a combination of a conventional prediction GP $g(\cdot)$ with an SDE flow GP $\f(\cdot)$ fully parameterised by $\Z,\U$ as well as kernel parameters $\theta$. We turn to variational inference to estimate posterior approximations $q(\Uf)$ and $q(\u_g)$ for both models.

Exact inference of Gaussian processes has a limiting complexity of $\mathcal{O}(N^3)$. Instead, we apply stochastic variational inference (SVI) \citep{hensman2015}, which has been demonstrated to scale GP's up to a billion data points \citep{salimbeni2017}. We here summarise the SVI procedure following \citet{hensman2015}.

We start with the joint density of a single path and prediction of the augmented system
\begin{align}
    &p(\y, \g, \ug, \XT, \f,\Uf )= \underbrace{p(\y | \g)}_{\text{likelihood}} \underbrace{p(\g | \ug, \XT)  p(\ug )}_{\text{GP prior of $g(\x)$}} \underbrace{p(\XT|\f )}_{\text{SDE}} \underbrace{p(\f | \Uf) p(\Uf)}_{\text{GP prior of $\f(\x)$}}. \label{eq:app_joint}
\end{align}

where we have augmented the predictor function $g$ with $M$ inducing locations $\Zg = (\z_{g1}, \ldots, \z_{gM})$ with associated inducing function values $g(\z) = u$ in a vector $\ug = (u_{g1}, \ldots, u_{gM})^T \in \R^M$ with a GP prior. The conditional distribution is \citep{titsias2009}
\begin{align}
p(\g | \ug, \XT) &= \N(\g | \Q_T \ug, \K_{\XT \XT} - \Q_T \K_{\Zg \Zg} \Q_T^T) \\
    p(\ug ) &= \N(\ug | \0, \K_{\Zg \Zg}),
\end{align}
where we denote $\Q_T = \K_{\XT \Zg} \K_{\Zg \Zg}^{-1}$. 

Similarly, the warping function $\f$ is augmented with inducing variables $\mathbf{u}_{\f d} = (\u_1^f(d)^T, \ldots, \u_M^f(d))$ and inducing locations $\Zf_d = (\z_1^f(d), \ldots, \z_M^f(d))^T$.

\begin{align}
    p(\Uf) &= \prod_{d=1}^D \N( \mathbf{u}_{\f d} | \0, \K_{\Zf_d \Zf_d}),
\end{align}

The joint distribution \eqref{eq:app_joint} contains the likelihood term, the two GP priors, and the SDE term $p(\XT | \f)$ representing the Euler-Maruyama paths of the dataset.
\begin{align}
    d\x_t &= \bmu(\x_t) dt + \sqrt{\Sigma(\x_t)} dW_t\\
    \bmu(\x_t) &= \bR \Uf \\
    \bS(\x_t) &= \K_{\x \x} - \bR \K_{\Zf \Zf} \bR
\end{align}

We consider optimizing the marginal likelihood
\begin{align}
\log p(\y) &= \log \E_{ p(\g|\XT)p(\XT) } p(\y | \g), \label{eq:app_mll} \\
p(\g|\XT) &= \int p(\g|\ug,\XT) p(\ug) d\ug\\
p(\XT) &= \iint p(\XT | \f ) p(\f | \Uf) p(\Uf) d\f d\Uf,
\end{align}
with no tractable solution due to the FPK state distribution $p(\XT)$. 

A variational lower bound for the evidence \eqref{eq:app_mll} without the state distributions has already been considered by \citet{hensman2015}. We propose to include the state distributions by simulating Monte Carlo state trajectories.

We propose a complete variational posterior approximation over both $\f$ and $g$,
\begin{align}
q(\g, \ug, \XT, \f, \Uf) &= p(\g | \ug,\XT) q(\ug) p(\XT|\f) p(\f | \Uf) q(\Uf) \\
q(\ug) &= \N(\ug | \mg, \Sg) \\ 
q(\Uf) &= \prod_{d=1}^D \N( \mathbf{u}_{\f d} | \mathbf{m}_{\f d}, \mathbf{S}_{\f d}),
\end{align}
where $\Mf = ( \mathbf{m}_{\mathbf{f} 1}, \ldots, \mathbf{m}_{\mathbf{f} D})$ and $\Sf = (\mathbf{S}_{\f 1}, \ldots, \mathbf{S}_{\f D})$ collect the dimension-wise inducing parameters. We continue by marginalizing out inducing variables $\ug$ and $\Uf$ from the above joint distribution arriving at the joint variational posterior
\begin{align}
q(\g, \XT, \f) &= q(\g | \XT) p(\XT | \f) q(\f),
\end{align}
where
\begin{align}
    q(\g | \XT) &= \int p( \g | \ug, \XT) q(\ug) d \ug \\
    &= \N( \g |\Q_T \mg, \K_{\XT \XT} + \Q_T(\Sg - \K_{\Zg \Zg}) \Q_T^T) \\
    q(\f) &= \int p( \f | \Uf) q(\Uf) d \Uf \\
    &= \N( \f |\bmu_q,\bS_q) \notag \\
    \bmu_q &= \Qf \Mf \\
    \bS_q  &= \K_{\X \X} + \Qf (\Sf - \K_{\Zf \Zf}) \Qf^T
\end{align}
where $\Qf = \K_{\X \Zf} \K_{\Zf \Zf}^{-1}$. We plug the derived variational posterior drift  $\bmu_q$ and diffusion $\bS_q$ estimates to the SDE to arrive at the final \emph{variational SDE flow}
\begin{align}
    d\x_t &= \bmu_q(\x_t) dt + \sqrt{\bS_q(\x_t)} dW_t,
\end{align}
which conveniently encodes the variational approximation of $\f$.

Now the lower bound for our differential deep GP model can be written as
\begin{align}
    \log p(\y) &\ge \int q(\g, \ug, \XT, \f, \Uf) \log \frac{ p(\y, \g, \ug, \XT, \f,\Uf )}{ q(\g, \ug, \XT, \f, \Uf)} d\g  d\ug  d\XT  d\f  d\Uf \\
    &\ge \int p(\g | \ug,\XT) q(\ug) p(\XT|\f) p(\f | \Uf) q(\Uf) \log \frac{p(\y | \g)p(\ug )p(\Uf)}{q(\ug)q(\Uf) } d\g  d\ug  d\XT  d\f  d\Uf \\
    &\ge \int q(\g |\XT) q(\XT) \log p(\y | \g)d\g  d\XT - \KL[ q(\ug) || p(\ug)] - \KL[ q(\Uf) || p(\Uf)]\\
    &\ge \sum_{i=1}^N \bigg\{ \frac{1}{S} \sum_{s=1}^S \underbrace{\E_{q( \g | \x_{i,T}^{(s)}) } \log p(y_i | g_i)}_{\text{variational expected likelihood}} - \underbrace{\KL[ q(\ug) || p(\ug)]}_{\text{$g(x)$ prior divergence}} - \underbrace{\KL[ q(\Uf) || p(\Uf)]}_{\text{$\f(x)$ prior divergence}} \bigg\}. \label{eq:app_elbof}
\end{align}
\newpage
\subsection*{B. Regression and classification benchmarks}

\begin{table*}[th]
\resizebox{1.00\textwidth}{!}{
\begin{tabular}{ l r  cc cc cc cc }
\toprule
& & boston & energy & concrete & wine\_red & kin8mn & power & naval & protein \\
\cmidrule(lr){3-10}
& $N$ & 506 & 768 & 1,030 & 1,599 & 8,192 & 9,568 & 11,934 & 45,730  \\
& $D$ & 13 & 8 & 8 & 22 & 8 & 4 & 26 & 9  \\
\midrule
Linear && 4.24(0.16) & 2.88(0.05) & 10.54(0.13) & 0.65(0.01) & 0.20(0.00) & 4.51(0.03) & 0.01(0.00) & 5.21(0.02)  \\
\cmidrule(lr){1-10}
BNN & $L=2$          & 3.01(0.18) & 1.80(0.05) & 5.67(0.09) & 0.64(0.01) & 0.10(0.00) & 4.12(0.03) & 0.01(0.00) & 4.73(0.01)  \\
\cmidrule(lr){1-10}
\multirow{2}{*}{Sparse GP}& $M=100$  & 2.87(0.15) & 0.78(0.02) & 5.97(0.11) & 0.63(0.01) & 0.09(0.00) & 3.91(0.03) & \textbf{0.00}(0.00) & 4.43(0.03)  \\
& $M=500$  & 2.73(0.12) & \textbf{0.47}(0.02) & 5.53(0.12) & \textbf{0.62}(0.01) & 0.08(0.00) & 3.79(0.03) &\textbf{0.00}(0.00) & 4.10(0.03)  \\
\cmidrule(lr){1-10}
\multirow{4}{*}{\shortstack{Deep GP \\ $M=100$}} & $L=2$      & 2.90(0.17) & \textbf{0.47}(0.01) & 5.61(0.10) & 0.63(0.01) & \textbf{0.06}(0.00) & 3.79(0.03) & \textbf{0.00}(0.00) & 4.00(0.03)  \\
& $L=3$      & 2.93(0.16) & 0.48(0.01) & 5.64(0.10) & 0.63(0.01) & \textbf{0.06}(0.00) & 3.73(0.04) & \textbf{0.00}(0.00) & \textbf{3.81}(0.04)  \\
& $L=4$      & 2.90(0.15) & 0.48(0.01) & 5.68(0.10) & 0.63(0.01) & \textbf{0.06}(0.00) & 3.71(0.04) & \textbf{0.00}(0.00) & \textbf{3.74}(0.04)  \\
& $L=5$      & 2.92(0.17) & \textbf{0.47}(0.01) & 5.65(0.10) & 0.63(0.01) & \textbf{0.06}(0.00) & 3.68(0.03) & \textbf{0.00}(0.00) & \textbf{3.72}(0.04)  \\
\cmidrule(lr){1-10}
\multirow{5}{*}{\shortstack{DiffGP \\ $M=100$}}
& $T= 1.0$ &2.80(0.13) & 0.49(0.02) & \textbf{5.32}(0.10) & 0.63(0.01) & \textbf{0.06}(0.00) & 3.76(0.03) & \textbf{0.00}(0.00) & 4.04(0.04) \\ 
& $T= 2.0$ & \textbf{2.68}(0.10) & 0.48(0.02) & \textbf{4.96}(0.09) & 0.63(0.01) & \textbf{0.06}(0.00) & 3.72(0.03) & \textbf{0.00}(0.00) & 4.00(0.04) \\
& $T= 3.0$ & \textbf{2.69}(0.14) & \textbf{0.47}(0.02)& \textbf{4.76}(0.12) & 0.63(0.01) & \textbf{0.06}(0.00) & 3.68(0.03) & \textbf{0.00}(0.00) & 3.92(0.04)  \\
& $T= 4.0$ & \textbf{2.67}(0.13) & 0.49(0.02) & \textbf{4.65}(0.12) & 0.63(0.01) & \textbf{0.06}(0.00) & \textbf{3.66}(0.03) & \textbf{0.00}(0.00) & 3.89(0.04) \\
& $T= 5.0$ & \textbf{2.58}(0.12) & 0.50(0.02) & \textbf{4.56}(0.12) & 0.63(0.01) & \textbf{0.06}(0.00) & \textbf{3.65}(0.03) & \textbf{0.00}(0.00) & 3.87(0.04)  \\
\bottomrule
\end{tabular}
}
\caption{Test RMSE values of 8 benchmark datasets (reproduced from from Salimbeni \& Deisenroth 2017). Uses random 90\% / 10\% training and test splits, repeated 20 times. }
\end{table*}

\begin{table*}[th]
\resizebox{1.00\textwidth}{!}{
\begin{tabular}{ l r  cc cc cc cc }
\toprule
& & boston & energy & concrete & wine\_red & kin8mn & power & naval & protein \\
\cmidrule(lr){3-10}
& $N$ & 506 & 768 & 1,030 & 1,599 & 8,192 & 9,568 & 11,934 & 45,730  \\
& $D$ & 13 & 8 & 8 & 22 & 8 & 4 & 26 & 9  \\
\midrule
Linear && -2.89(0.03) &-2.48(0.02)& -3.78(0.01) & -0.99(0.01) & 0.18(0.01) & -2.93(0.01) & 3.73(0.00) & -3.07(0.00)  \\
\cmidrule(lr){1-10}
BNN & $L=2$          & -2.57(0.09) & -2.04(0.02) & -3.16(0.02) & -0.97(0.01) & 0.90(0.01) &-2.84(0.01) &3.73(0.01) & -2.97(0.00)  \\
\cmidrule(lr){1-10}
\multirow{2}{*}{Sparse GP}& $M=100$  & -2.47(0.05) & -1.29(0.02) & -3.18(0.02)
 & -0.95(0.01) & 0.63(0.01)& -2.75(0.01) & 6.57(0.15) & -2.91(0.00) \\
& $M=500$  & -2.40(0.07) & \textbf{-0.63}(0.03)& -3.09(0.02) &  \textbf{-0.93}(0.01) & 1.15(0.00) &-2.75(0.01) & \textbf{7.01}(0.05) & -2.83(0.00)  \\
\cmidrule(lr){1-10}
\multirow{4}{*}{\shortstack{Deep GP \\ $M=100$}} & $L=2$      &-2.47(0.05) & -0.73(0.02) & -3.12(0.01) & -0.95(0.01)&1.34(0.01) & -2.75(0.01) &  6.76(0.19) & -2.81(0.00)  \\
& $L=3$      & -2.49(0.05) & -0.75(0.02) & -3.13(0.01) & -0.95(0.01)& 1.37(0.01) & -2.74(0.01) & 6.62(0.18) & \textbf{-2.75}(0.00)  \\
& $L=4$      &-2.48(0.05) & -0.76(0.02) & -3.14(9
.01) & -0.95(0.01)& \textbf{1.38}(0.01) & -2.74(0.01) & 6.61(0.17) & \textbf{-2.73}(0.00)  \\
& $L=5$      &-2.49(0.05) &-0.74(0.02) & -3.13(0.01) & -0.95(0.01)& \textbf{1.38}(0.01) &-2.73(0.01) & 6.41(0.28) & \textbf{-2.71}(0.00)  \\
\cmidrule(lr){1-10}
\multirow{5}{*}{\shortstack{DiffGP \\ $M=100$}}
&$T= 1.0$ &\textbf{-2.36}(0.05)& -0.65(0.03)& \textbf{-3.05}(0.02)& -0.96(0.01)& 1.36(0.01)& -2.75(0.01)& 6.58(0.02)& -2.79(0.04)\\ 
&$T= 2.0$ & \textbf{-2.32}(0.04)& \textbf{-0.63}(0.03)& \textbf{-2.96}(0.02)& -0.97(0.02)& 1.37(0.00)& -2.74(0.01)& 6.26(0.03)& -2.78(0.04)\\ 
&$T= 3.0$ &\textbf{-2.31}(0.05)& \textbf{-0.63}(0.02)& \textbf{-2.93}(0.04)& -0.97(0.02)& 1.37(0.01)& \textbf{-2.72}(0.01)& 6.00(0.03)& -2.79(0.00)\\ 
&$T= 4.0$ &\textbf{-2.33}(0.06)& -0.65(0.02)& \textbf{-2.91}(0.04)& -0.98(0.02)& 1.37(0.01)& \textbf{-2.72}(0.01)& 5.86(0.02)& -2.78(0.00)\\ 
&$T= 5.0$ &\textbf{-2.30}(0.05)& -0.66(0.02)& \textbf{-2.90}(0.05)& -0.98(0.02)& 1.36(0.01)& \textbf{-2.72}(0.01)& 5.78(0.02)& -2.77(0.00)\\ 
\bottomrule
\end{tabular}
}
\caption{Test log likelihood values of 8 benchmark datasets (reproduced from from Salimbeni \& Deisenroth 2017)}
\end{table*}

\begin{table}[th]
\centering
\resizebox{0.45\columnwidth}{!}{
\begin{tabular}{ l r c c }
\toprule
 & & SUSY & HIGGS \\
\cmidrule(lr){3-4}
 & $N$ & 5,500,000 & 11,000,000\\
 & $D$ & 18 & 28 \\
 \midrule
\midrule
DNN & & 0.876 & \textbf{0.885}\\
\multirow{2}{*}{Sparse GP} & $M=100$ & 0.875 & 0.785 \\
& $M=500$ & 0.876 & 0.794 \\
\midrule
\multirow{4}{*}{\shortstack{Deep GP \\ $M=100$}} & $L=2$      & 0.877 & 0.830 \\
& $L=3$      & 0.877 & 0.837 \\
& $L=4$      & 0.877 & 0.841 \\
& $L=5$      & 0.877 & \textbf{0.846} \\
\midrule
\multirow{2}{*}{\shortstack{DiffGP \\ $M=100$}} 
& $t=1.0$ & \textbf{0.878} & 0.840 \\
& $t=3.0$ & \textbf{0.878} & 0.841 \\
& $t=5.0$ & \textbf{0.878} & 0.842 \\
\midrule
\multirow{2}{*}{\shortstack{DiffGP Temporal\\ $M_s =100$ \\$M_t=3$}} 
\\
& $t=5.0$ & \textbf{0.878} & \textbf{0.846} \\
\\
\bottomrule
\end{tabular}
}
\caption{Test AUC values for large-scale classification datasets. Uses random 90\% / 10\% training and test splits.}
\end{table}

\end{document}